\documentclass{article}

\PassOptionsToPackage{sort,numbers}{natbib}

\usepackage[dblblindworkshop, final]{neurips_2025}

\usepackage[utf8]{inputenc} %
\usepackage[T1]{fontenc}    %
\usepackage{hyperref}       %
\usepackage{url}            %
\usepackage{booktabs}       %
\usepackage{amsfonts}       %
\usepackage{nicefrac}       %
\usepackage{microtype}      %
\usepackage{xcolor}         %
\usepackage{graphicx}
\usepackage{csquotes}

\usepackage{todonotes}

\title{Efficiency Will Not Lead to Sustainable Reasoning AI}
\workshoptitle{Rethinking AI}

\author{
  Philipp Wiesner$^1$, Daniel W. O'Neill$^2$, Francesca Larosa$^3$, Odej Kao$^1$ \\
  $^1$TU Berlin, Germany \\
  $^2$Universitat de Barcelona, Spain \\
  $^3$KTH Royal Institute of Technology, Sweden \\
  \texttt{\{wiesner,odej.kao\}@tu-berlin.de, oneill@ub.edu, larosa@kth.se} \\
}

\begin{document}

\maketitle

\begin{abstract}
AI research is increasingly moving toward complex problem solving, where models are optimized not only for pattern recognition but for multi-step reasoning. 
Historically, computing's global energy footprint has been stabilized by sustained efficiency gains and natural saturation thresholds in demand. 
But as efficiency improvements are approaching physical limits, emerging reasoning AI lacks comparable saturation points: performance is no longer limited by the amount of available training data but continues to scale with exponential compute investments in both training and inference.
This paper argues that efficiency alone will not lead to sustainable reasoning AI and discusses research and policy directions to embed explicit limits into the optimization and governance of such systems.

\end{abstract}

\section{Introduction}

For decades, the demand for computing has expanded rapidly without a proportional rise in energy use~\cite{masanet2020}. 
Successive technological waves—search, cloud computing, social media, or on-demand video streaming—have driven exponential increases in computational demand, yet the total electricity use of data centers has remained relatively stable. 
Between 2010 and 2018, for example, global data center electricity consumption increased by only about 6\% (to roughly 205 TWh, around 1\% of global electricity use), while the amount of computation performed grew by 550\%~\cite{masanet2020}.
This stability was achieved through continuous improvements in hardware and software efficiency~\cite{bashir2023hotair,Muralidhar2020Energy}, and each wave of expansion eventually slowed as markets matured and user needs saturated.

This apparent balance between demand and efficiency created the expectation that technological progress naturally offsets growth in demand. 
However, the emerging trend toward reasoning models might break this pattern. 
Unlike earlier models, reasoning AI is less constrained by the availability of training data and can continue to improve primarily through additional computation, both during inference (via multi-step reasoning~\cite{wei2022chain,wang2023selfconsistency,yao2023tree,besta2024graph}) and training (usually via reinforcement learning~\cite{ouyang2022training,openai2024,google2024gemini,deepseekai2025deepseekr1incentivizingreasoningcapability}).
Exponential computational investment can enable deeper reasoning, higher accuracy, and broader generalization, developments often viewed as steps toward more general or even “superintelligent” AI systems.

Reasoning AI is ultimately constrained by efficiency itself: any gains can be \emph{immediately} reinvested. 
While such rebound effects have long been characteristic of computing, we argue that in the age of reasoning AI, their scale has reached a point where improvements in hardware or software efficiency can no longer be viewed as inherently contributing to sustainability.

\section{Advances in Structured and Reinforced Reasoning}

This section surveys recent developments in reasoning AI and examines why they could result in escalating, and in some cases unbounded, computational demand for both inference and training.

\paragraph{Structured prompting}

Early transformer-based models demonstrated impressive linguistic fluency but limited capacity for logical inference, leading to their description as \enquote{stochastic parrots}~\cite{bender2021parrots}. 
This changed with the introduction of \emph{Chain-of-Thought}~\cite{wei2022chain,wang2023selfconsistency} prompting, which conditions models to generate intermediate reasoning steps before producing a final answer. 
This technique greatly improved performance on arithmetic and symbolic reasoning tasks, suggesting that pretrained models already contain latent reasoning abilities that surface with longer inference paths.

Subsequent work extended this paradigm toward richer, non-linear reasoning structures. 
For example, \emph{Tree-of-Thought}~\cite{yao2023tree, xie2023selfevaluation, long2023large} models reasoning as a form of guided search, in which multiple partial reasoning paths can be expanded, evaluated, and pruned during inference. 
\emph{Graph-of-Thought}~\cite{besta2024graph, lei2023boosting} reasoning generalizes this concept by allowing arbitrary directed dependencies among intermediate thoughts, enabling branching, recombination, and iterative self-correction. 
These structured prompting approaches transform inference from a simple forward pass into an adaptive exploration process, with accuracy improving monotonically as computation increases.

\paragraph{Reinforced reasoning}

Building on these advances, researchers began to complement structured prompting with reinforcement learning~(RL) to refine reasoning behavior. 
In RL, a model interacts with an environment, receives feedback, and updates its policy to maximize a reward signal. 
Applied to language models, approaches like InstructGPT~\cite{ouyang2022training} applied RL based on human feedback, training a reward model to prefer outputs that are coherent, useful, or correct.
\emph{Sparrow}~\cite{glaese2022improving} extended this idea to factuality and rule adherence. 
More recently, OpenAI’s o1~\cite{openai2024}, Google DeepMind’s Gemini~\cite{google2024gemini}, or DeepSeek’s R1~\cite{deepseekai2025deepseekr1incentivizingreasoningcapability} have applied RL directly to reasoning, rewarding models that deliberate longer, verify intermediate steps, or self-correct errors. 

Through feedback-driven training, reasoning becomes a self-optimizing process: models can, in principle, continue improving as long as additional compute and reward signals are available.
To scale this feedback beyond human supervision, researchers increasingly employ \emph{LLMs as judges}, where a secondary model evaluates the reasoning quality or correctness of another~\cite{gu2024survey, bai2023rlaif}. 
This automated feedback enables large-scale reinforcement learning without human annotators, allowing reasoning models to refine themselves through iterative self-assessment.
This approach is, for example, increasingly used in mathematics~\cite{deepmind2024imo, lin2025goedelproverv2, ren2025deepseekproverv2, wang2025kimina_preview}, where rewards can be precisely defined by proof correctness, and reasoning LLMs might soon approach problems beyond human reach.

\section{Why Efficiency Is Not Enough}

Efficiency has long been considered the main lever for \enquote{green} or \enquote{sustainable} computing, yet its stabilizing effect is increasingly illusory.
Across the computing stack, each new generation of chips, data center infrastructure, and software optimization has reduced the energy required for a given task.
However, these improvements have rarely led to reduced overall energy use.
Instead, cheaper and more efficient computation has consistently stimulated new applications, larger models, and more intensive use, a dynamic known as the \emph{rebound effect} or \emph{Jevons’ paradox}~\cite{alcott2005jevons, luccioni2025reboundAI}.

\paragraph{Rebound effects} Jevons observed in 1865 that technological improvements in coal engine efficiency did not reduce total coal consumption but instead increased it by making coal more economically attractive and opening new applications~\cite{alcott2005jevons}. 
For a direct rebound effect to occur, two conditions are required: (1) an improvement in efficiency, and (2) elastic demand for the product in question---in this case, computational power. 
For much of the ICT sector’s history, the second condition was constrained, as most workloads eventually reached market or human attention limits.
Under these circumstances, efficiency gains could keep total energy use roughly constant.
However, as outlined earlier, reasoning models exhibit virtually unbounded demand: as long as compute yields better performance, there is no natural saturation point. 
Even if hardware becomes another 10× more efficient and software 1000×, these gains would likely be reinvested into deeper reasoning and larger deployments rather than reducing energy and resource use.

\paragraph{Diminishing returns}

For decades, improvements in hardware and infrastructure efficiency have been the primary factor keeping the energy footprint of computing under control.
However, physical limitations---such as the fundamental energy limits of computation~\cite{bennett1985fundamental}, power density constraints~\cite{Chang2010Practical}, and the end of Moore's Law~\cite{theis2017endofmooreslaw}---mean that further gains in hardware efficiency are becoming increasingly difficult and expensive to achieve~\cite{bashir2023hotair, Muralidhar2020Energy}.
Data centers already operate near their thermodynamic and engineering limits: the power usage effectiveness of hyperscale data centers has plateaued around 1.1~\cite{google2025pue}, leaving little room for further reductions in overhead energy use~\cite{thompson2021}.
While new cooling concepts and workload management strategies can still yield incremental benefits, the large efficiency improvements that previously offset exponential growth are no longer realistic.

Software remains a major source of potential efficiency, through advances in algorithms, compilers, and model architectures.
However, these improvements increasingly operate within tight bounds and are typically absorbed by the rebound effects described above.
Every gain in model compression, quantization, or hardware acceleration lowers the cost per inference or training step and thus increases the incentive to perform more of them.
When both hardware and infrastructure efficiency approach their physical limits, and software improvements merely accelerate utilization, total energy consumption begins to scale more directly with computational demand.

\paragraph{Reasoning AI lacks saturation thresholds}

In the current \enquote{Era of Human Data}~\cite{silver_sutton2025era_of_experience}, model performance was bounded by the quantity and quality of human-generated content, or as Ilya Sutskever said, \enquote{there’s only one internet}~\cite{sutskever2024}.
Once large language models had absorbed most high-quality text, further scaling produced diminishing returns, motivating a shift toward smaller, more efficient architectures and techniques such as pruning, quantization, and low-rank adaptation.
In this setting, efficiency directly translated into lower costs and emissions because the underlying demand for computation was finite.

In the \enquote{Era of Experience}~\cite{silver_sutton2025era_of_experience}, as put by Silver and Sutton, reasoning AI removes these constraints.
These models are not limited by fixed datasets or human supervision but can generate and evaluate their own training data through reinforcement learning and self-assessment.
Each additional unit of compute can yield longer chains of thought, deeper search trees, or more refined self-corrections.
This property effectively eliminates the saturation thresholds that once contained computational demand.

\section{Accelerating Compute Expansion}%

The past year has seen an unprecedented acceleration in the construction of AI infrastructure, suggesting that computational capacity is entering a new phase of growth rather than nearing saturation.

For example, xAI claimed to have built \enquote{Colossus}, described as the world's largest AI supercomputer, in just 122 days, and to have doubled the cluster to 200,000 GPUs within weeks~\cite{fortune2024elonmusk}. 
The facility is reportedly powered primarily by methane gas turbines, many of which are believed to operate without full permits, leading to substantial local air pollution~\cite{xai_memphis2025guardian} and community backlash~\cite{community2025nyt}.
Nevertheless, xAI recently raised about \$20~billion to fund further data center expansions~\cite{xai20b2025reuters}.
Meta has announced multiple multi-gigawatt data centers as part of their \enquote{titan cluster}, and Zuckerberg pledged to invest \enquote{hundreds of billions} for superintelligence infrastructure~\cite{meta_supercluster}. 
Recently, Sam Altman stated the current goal of OpenAI is to \enquote{create a factory that can produce a gigawatt of new AI infrastructure every week}~\cite{altman_abundant}. 

These moves and statements reveal three dynamics: (1) companies are targeting gigawatt-scale power budgets, not incremental upgrades; (2) buildout timelines are shrinking from years to months; and (3) the capital commitment is enormous, suggesting that compute availability has become the primary competitive constraint.

Together, these observations offer early evidence that AI compute is entering a new phase of expansion and they call for a reflection about the energy efficiency debate.
First, many of these supercomputers are expected to rely primarily on coal or gas~\cite{Robinson2025DataCenterCoal, DeChant2025GasPowerMeta}, directly coupling their expansion to rising emissions. Given the current AI arms race, there is a real risk that companies and nations will prioritize computational advantage over sustainability.
Second, there is a growing mismatch between the pace of compute expansion and investments in supporting infrastructure. As the technology becomes more accessible, the risk of demand-driven resource overshoot increases.
Investments in grid resilience and low-carbon generation must therefore be made in anticipation, not in response, to the rapid adoption of AI. 
Acquisitions of nuclear power assets and efforts to deploy low-carbon technologies in data centers~\cite{IEA2025Nuclear} demonstrate that companies recognize the need for reliable, continuous power.
Without coordinated planning between AI infrastructure development and the energy system, the resulting strain could undermine grid stability and the credibility of the industry’s sustainability commitments, as well as the promises made to investors.

\section{Future Research Directions and Policy Implications}

Ensuring that progress in reasoning AI remains compatible with planetary limits requires coordinated advances in measurement and accounting, the detection and mitigation of rebound effects, and the development of better governance frameworks.

\paragraph{Accounting and operational carbon efficiency}

Unified and verifiable carbon accounting is fundamental to any effective policy or governance framework. 
Achieving this requires consistent reporting methodologies of energy use and carbon emissions across hardware platforms, model architectures, and both training and inference strategies.
In parallel, we must establish robust and comparable metrics to assess real progress toward sustainability, capturing not only relative efficiency but also absolute and life-cycle impacts. 
Such information supports policy design and can be used to optimize AI workload scheduling~\cite{bashir2024climate}.
By shifting computation across time or regions to better align with renewable energy availability---an approach known as carbon-aware computing~\cite{radovanovic2021carbonaware}---we can reduce the operational footprint of AI, for example, by adjusting the energy spent per inference request based on grid conditions~\cite{li2024sprout, wiesner2025quality}. However, carbon efficiency alone does not guarantee sustainability, as it can still lead to rebound effects through expanded use of cheaper or cleaner computation.

\paragraph{Embedding limits into optimization}

As reasoning systems become increasingly autonomous, we must develop mechanisms that constrain or audit their capacity for recursive self-improvement within fixed energy or carbon budgets.
For example, RL frameworks could incorporate environmental externalities into their reward functions, enabling models to learn not only accuracy but also efficiency in computation and energy use. 
Compute governance APIs could operationalize these constraints in practice, allowing autonomous reasoning systems to self-train or self-evaluate only within verifiable resource limits.
Such interfaces would make energy-aware and carbon-aware optimization an enforceable property of intelligent systems.

\paragraph{Enabling policy responses to rebound effects}

Two policy responses effectively limit rebound effects~\cite{alcott2010impact}: First, \emph{caps on resource use} establish physically defined limits on energy or material consumption, applied either upstream (at production) or downstream (at consumption). For AI systems, this could mean restricting the total energy available to data centers for training and inference or introducing personal and institutional \enquote{compute budgets.} Such measures ensure sustainability by definition, as total resource use cannot exceed the cap. Second, \emph{Pigouvian taxes} achieve similar outcomes through price mechanisms rather than quantity restrictions. By internalizing environmental costs, sufficiently high taxes make resource-intensive AI applications economically viable only when they deliver commensurate social value. 
Both caps and taxes, however, must be informed by accurate and transparent data on the energy use, carbon intensity, and life-cycle impacts of computing systems---information that computer scientists must provide and standardize.

\paragraph{Prioritizing applications that improve human wellbeing and planetary stewardship}

A complementary approach involves restricting computationally intensive reasoning AI to applications that demonstrably advance human wellbeing. Rather than allowing market forces alone to determine AI deployment, policy could require justification against established sustainability frameworks such as the Sustainable Development Goals \cite{vinuesa2020sdg} or the Doughnut of social and planetary boundaries \cite{raworth2017doughnut}. AI development can also be steered toward Earth-aligned applications that prioritize climate-compatible activities to support adaptation and mitigation efforts~\cite{Gaffney2025EarthAlignment}. For example, reasoning models could be applied to accelerate cancer research, optimize renewable energy systems, or improve climate modeling, while restricting their use for applications with limited social value such as holiday planning, personalized advertising, or entertainment. 
This requires a periodic assessment of the environmental costs and societal benefits of AI systems, including their direct, indirect, and systemic impacts.

\section{Conclusion}

In this paper, we argue that efficiency alone cannot contain the fundamentally unbounded computational and energy demand of reasoning AI. 
Sustainable progress therefore requires frameworks that make this expansion both measurable and governable. Future research should focus on developing methods to anticipate and mitigate the resource footprint of reasoning systems, ensuring their evolution remains compatible with planetary boundaries.

\begin{ack}
DWO acknowledges funding by the European Union in the framework of the Horizon Europe Research and Innovation Programme under grant agreement number 101137914 (MAPS: ``Models, Assessment, and Policies for Sustainability'').

FL acknowledges the LIBRA project, funded by the European Union's Horizon Europe research and innovation programme under the Marie Skłodowska-Curie grant agreement No. 101150729.
\end{ack}

{
\small

\bibliography{bibliography}
\bibliographystyle{plainurl}  %

}

\end{document}